# Theoretical Unification of the Fractured Aspects of Information

(Draft of a paper intended as a chapter in the forthcoming book Schroeder, M. J. & Hofkirchner, W. (eds.) Understanding Information and Its Role as a Tool: In Memory of Mark Burgin (in 2 Parts), Part I: Understanding Information: Theory and Foundations. World Scientific Publishing, Singapore, archived for the purpose of research cooperation. Please make references to the final version in this book.)

**Marcin J. Schroeder**

*Akita International University*
*Yuwa, Tsubakigawa, Akita, 010-1211 Japan*
*mjs@gl.aiu.ac.jp*

The article has as its main objective the identification of fundamental epistemological obstacles in the study of information related to unnecessary methodological assumptions and the demystification of popular beliefs in the fundamental divisions of the aspects of information that can be understood as Bachelardian rupture of epistemological obstacles. These general considerations are preceded by an overview of the motivations for the study of information and the role of the concept of information in the conceptualization of intelligence, complexity, and consciousness justifying the need for a sufficiently general perspective in the study of information, and are followed at the end of the article by a brief exposition of an example of a possible application in the development of the unified theory of information free from unnecessary divisions and claims of superiority of the existing preferences in methodology. The reference to Gaston Bachelard and his ideas of epistemological obstacles and epistemological ruptures seems highly appropriate for the reflection on the development of information study, in particular in the context of obstacles such as the absence of semantics of information, negligence of its structural analysis, separation of its digital and analog forms, and misguided use of mathematics.

*This work is dedicated to the memory of Mark Burgin who contributed to the study of information and in particular to its theoretical development not only through the writing and publishing of many works of fundamental importance for the subject but also by his work on editing books, organizing conferences, and mentoring younger researchers.*

## 1. Introduction

The myth of possible or even necessary separation of the syntactic and semantic aspects of thought and language persists and spills over to the study of information where it is frequently conflated with the belief that the quantitative analysis of information eliminates the need for a seemingly inferior inquiry of its qualitative characteristics such as structure or modes of existence.

The exclusive focus on the quantitative methodology in the study of information is a product of the much wider tendency in the scientific methodology of many domains of inquiry produced by the illusion of precision and easy understanding of results expressed with the use of numbers. The focus on numbers brings another confusion into the study of information regarding the distinction between analog and digital types of information and information processing.

Yet another result of the belief in the distinction and superiority of the quantitative methodology is the conclusion drawn from the fact that probability theory and statistics have so broad and successful applications in a very large variety of disciplines that the probabilistic description of the concept of information explains its omnipresence in the inquiry of reality. Surprisingly, the calls for reversing the roles of information and probability and the use of the concept of information as a foundation for the study of probability made by several distinguished mathematicians of the $20^{th}$ century did not receive a sufficient response.

This article has as its main objectives the identification of such fundamental epistemological obstacles in the study of information related to unnecessary, hidden methodological assumptions and the demystification of popular beliefs in the fundamental divisions of the aspects of information in the hope that they bring Bachelardian epistemological unifying rupture. These general considerations are followed by a brief exposition of an example of the unified theory of information free from unnecessary divisions and claims of superiority of the existing preferences in methodology.

The reference to Gaston Bachelard and his ideas of epistemological obstacles and epistemological ruptures seems highly appropriate for the reflection on the development of information study. Bachelard was aware of the unavoidable obstacles created by our intuitive, common-sense conceptual and methodological frameworks which have to be identified and finally eliminated to achieve scientific progress [Bachelard, 1986; Tiles, 1984]. Bachelardian rupture is desired particularly in the context of obstacles such as the absence of unrestricted to particular contexts semantics of information, negligence of its structural analysis, and separation of its digital and analog forms.

The study of the epistemological obstacles in the study of information is preceded in this paper by an exposition of the reasons why their elimination is of great importance not just to satisfy researchers' curiosity. The main argument for revisiting the methodology of information studies and for maintaining their high level of generality is the relation of information to several other fundamental, but insufficiently conceptualized ideas such as intelligence, complexity, and consciousness which makes information a suitable defining concept.

Recent hot discussions of the danger of lost control over information technologies bring the subject to the attention of the global audience. All these three ideas of intelligence, complexity, and consciousness are at the center of the discussions, despite their vague, common sense understanding. The study of information is of its own interest, but its role as a foundation for the studies of these even less understood ideas makes it a prerequisite for solving one of the greatest challenges for humanity. It is argued in the following text that the only way to prevent the loss of control over information technology is not by blind, uninformed preventive legal regulation but by raising the understanding of the central concept of information.

## 2. Urgent Motivation for the Study of Information

### 2.1. *Understanding Generative Artificial Intelligence*

The concept of information, or rather in the absence of a commonly accepted definition its phantom invoked in discourses to create mutual understanding, is haunting the intellectual discussions on virtually every subject. Information is in the ghostly company of other formidable phantoms of concepts such as consciousness, complexity, intelligence, volition, computation, and life. All of them, or rather their elusiveness generate myths and anxiety about hidden dangers while the real danger, ignorance of their meaning and role, is in the open view.

Most recently the biggest splash has been made by the scepter of Artificial Intelligence whose name in the tradition of all taboos is fearfully expressed with the omnipresent acronym AI ("you know who", or in this case, "you know what"). Yet another source of anxiety expressed in media and on the internet is the possibility of encounters with extraterrestrial intelligence (ETI). AI seems dangerous as it can escape human rational control as if humans were in rational control of any large-scale phenomena. Of course, there are several well-known existential threats to humanity, such as climate change, non-sustainable use and management of natural resources, pandemics, nuclear weapons, and misuse of nuclear technology, etc. However, each of them is dangerous not because of its inevitability or the lack of knowledge or understanding of prevention but because of the rather irrational actions of humans or the lack of a coordinated effort. There is a common belief that if only humanity decides to be rational the threats would be overcome, so we do not have to worry about the danger.

The escape of AI systems from human rational control and encounters with ETI seem more threatening because AI can become more intelligent than humanity or humans and ETI would have been more intelligent if it could visit us from far away. It does not matter, at least for many of those who advocate the moratorium on AI research, that it is not clear what it means to be intelligent or why something intelligent out of human control is more dangerous than something that lacks or is deficient in intelligence (for instance a deranged, immoral, and power-greedy human who managed to take control of nuclear weapons through the corruption of political mechanisms).

The dangers of abuse, misuse, or loss of control of information technologies are real, but this applies to every type of technology (e.g. nuclear technologies). However, instead of calling for a moratorium on AI research before there is a sufficient legal regulation system, it is necessary to stimulate and support research on the concepts and ideas involved in the development of AI technology. Can anyone develop an effective legal regulation system for the developers of AI technologies to prevent all possible harm to humanity when so little is known or understood about the fundamental concepts of information studies?

The only way to prevent the harmful unexpected impact of technologies (all technologies, not just AI) is to create a legal requirement of sufficient investment in independent research on all aspects of their development, creation, and use. The requirement should apply not only to commercial developers of technologies but also to governmental agencies interested in their development. The less predictable the consequences of a given technology, the more fundamental and more intensive research should be mandated. In the case of AI technologies, it is almost impossible to make any predictions in the absence of sufficiently developed studies of consciousness, complexity, intelligence, volition, computation, and life. Thus, any organization engaging in the development of AI technologies should demonstrate investment in these studies. The lack of such investment could and should be used in the future as evidence for possible future liability. There should be no excuse for the insufficient knowledge of the possible consequences of technological innovation in the absence of documented sufficient investment in independent research not just on the subject of engineering, economic, or social aspects of its products, but on the more general task of understanding all phenomena involved in technological processes.

The first step in the direction of understanding AI technologies is the prevention of confusion proliferated by the everyday language of the news and commentaries on the subject. The expressions "AI can do..." or "AI becomes more intelligent than humans" suggest that AI is an entity or agent. There are important issues involved in understanding the words "artificial" and "intelligence", but even if we ignore them the expressions suggesting the uniform and independent ontological status of AI are misleading. Fortunately, this category error has been identified and criticized in the recent editorial in Nature Reviews Physics [Shevlin and Halina, 2019; Editorial, 2023].

"Anthropomorphic language is widespread in physics: masses 'feel' the gravitational potential, photons 'know' the state of their entangled partner and spins generally 'want' to align. [...]First, we will try to avoid at all costs the use of 'the AI/an AI' due to its unfortunate suggestion of agency. Instead, we will either change to 'the AI system/an AI system' or be very clear what we are talking about" [Editorial, 2023].

There are many reasons why this category error may have detrimental consequences for understanding the real reasons for concerns about AI. In the context of our paper, it makes an impression that AI is an existing and independent entity that can be considered in separation from the more fundamental study of information. There is no way to acquire knowledge and understanding of AI without a prior deep and extensive understanding of information and its involvement in phenomena related to consciousness, complexity, intelligence, volition, computation, and life. Of course, there is nothing wrong with using an informal abbreviation AI for the name of the entire complex of information technologies, but statements that AI can or cannot do something is an abuse of language.

AI systems are simply instances of devices designed by humans with the possible help of technological tools in which information dynamics is used to perform some actions. Thus far, the operation of such systems is controlled by human agents, but the control is declining. The devices of generative AI systems are designed to minimize the involvement of slow human agents. This follows the intentional and commonly accepted direction of technological progress which started two hundred years ago from the mechanization of work (elimination of the work of human or animal muscles) and was followed by automation (elimination of human control of machines. The difference in the design of the generative AI systems is in the elimination of meta-control. The automata used in manufacturing follow the process designed and controlled by human programmers. The generative AI systems increasingly act as "black boxes" whose operation is in principle known, but whose actual functioning is more and more autonomous based on the input not from particular human agents but from the data obtained from the internet. Thus, the control by humans is relinquished to the huge data reservoir on the internet that is "teaching" the system how to operate.

The question of whether generative AI systems are intelligent can be reformulated as to whether the internet (or any other dynamical data set) is intelligent. Another possibility is to look for the intelligence of generative AI systems in the ability to use the internet as a non-intelligent information resource consisting of discarded byproducts of human activities. This may be reassuring because these resources are human products and it seems that the exceeding of human capacities based on rather random human individual contributions is unlikely, but such optimism is unwarranted as information technology may detect patterns in the collective human activities that are beyond individual human comprehension. It seems a bizarre idea that any new great development in science (comparable to relativity theory or quantum mechanics in physics) could arise without human engagement from the patterns in the data stored on the internet when the generative AI system is trained on social media, but training on the archive of the entire scientific and philosophical heritage makes it more likely. Is the detection of patterns in collective knowledge sufficient for intelligence capable of creativity significantly exceeding human capacities? There is no answer to such questions about the intelligence of AI systems as long as we do not have a clear understanding of intelligence artificial or natural.

The next step is to consider the possibility that some AI systems may acquire the ability to build other AI systems (both as informational and natural/physical entities) and become autonomous natural devices. Here is the essence of the escape from human control. The danger is that some AI systems equipped with physical instruments that make them agents may acquire the ability to proliferate and act independently from human control and understanding. We already have examples of simpler systems that proliferate themselves (e.g. computer malware), but they still rely on devices created by humans and on human (usually unintentional) actions. But we know that self-reproducing physical automata are possible.

This possibility of escape is scary, but its threat is not new as the development of technologies generated concerns about autonomous devices several times in the last two centuries. However, is it so different from the issues related to invasive species in ecological systems or pandemics? In all cases, we need a better understanding of the dynamics of information in multiple contexts. The escape of a virus from a laboratory may be equally dangerous. Does it help that the virus has intelligence incomparable to human intelligence? The key point is that to prevent escapes of natural or artificial agents from human control we have first to acquire this control in the form of the knowledge of information dynamics.

## 2.2. *Can Artificial Intelligence Be Conscious?*

We could continue the search for multiple confusing and concerning aspects of AI and each time we will arrive at the same obstacle of the lack of understanding of the fundamental concepts related to information.

Can AI systems be conscious? The answer depends on how we understand consciousness, and whatever consciousness is, its understanding requires an explanation of how consciousness is related to information and its dynamics. The danger is in making assumptions based on popular common sense metaphors such as that the mind or brain is a type of computer or vice versa that computers are artificial minds or brains. Even more dangerous is the lack of distinctions between mind and brain. The popular analogy of software and hardware appeals to common sense, but it is based on the lack of knowledge of both, information mechanisms in the computer and in the brain.

The question about the possibility of conscious artificial intelligence systems is crucial for the consideration of their agency. When we talk about artificial intelligence and the possibility of systems independent from human control or comprehension, it is a legitimate question about not only their cognition but also conation. As was observed before, there are many systems (ecological, social, economic, cultural, etc.) that are independent of individual or collective human control. However, they are definitely devoid of purpose and their dynamics are governed by natural or social laws typically well known. They do not set or modify their goals and their mechanisms are driven by well-understood external forces. This gives hope to humanity for gaining control over them. The possibility of the intentional or unintentional construction of conscious artificial intelligence systems brings into consideration the danger of actual lost control. This requires some additional elements such as the capacity for self-consciousness, volition, and independent from human control normativity (the ability to set own values and goals). Only systems equipped with consciousness having these additional characteristics can compete with humanity. Otherwise, the main threat to humanity is humanity itself. However, is artificial consciousness possible?

There is no doubt that without substantial progress in the study of information, there will be no answers to the questions about the prevention of harm caused by present and future technologies. After all, the control of any artifacts comes not from watching and directing their work (technological progress was always generated by the interest in the elimination of both), but from the knowledge and understanding of the phenomena involved in their mechanisms allowing for the prediction of the outcomes of this work and prevention of their deviation from human goals and values. The expectation that the official moratorium on the research and development of AI can prevent future disasters is naive. Instead, there should be more support for the study of information going way beyond its technological aspects. The actions of the external support for the study of information should be informed and guided by the research done by the community dedicated to this study. This article is intended as a contribution to this goal.

## 3. The Role of Information in Understanding Intelligence, Complexity, and Consciousness

### 3.1. *Information and Intelligence*

Without claiming the achievement of the ultimate resolution of the still open issue of what intelligence is, I will use in this paper the concept of intelligence as a capacity to eliminate or decrease the complexity of information [Schroeder, 2020a].

The use of the concept of complexity in defining intelligence may generate an objection that I listed above both as poorly understood and insufficiently conceptualized. However, the study of complexity has acquired quite an advanced level in more specific contexts of inquiry (e.g. computing, dynamical systems). Moreover, complexity, no matter how defined, is a more general concept that can be applied at any level of abstraction to qualify arbitrary subjects of inquiries and actually is applied in a wide variety of contexts, while intelligence can characterize only systems capable of action, the action itself, or its outcome. It is easy to find examples of something complex that cannot be considered intelligent. On the other hand, everybody would agree that human intelligence is associated with the extreme complexity of the brain. This of course does not demonstrate that intelligence can be defined by complexity, but only that complexity cannot be defined by intelligence. Finally, complexity is frequently studied in the context of information. The next sub-section of this work is devoted to their relationship.

The difficulties in defining intelligence are well-known. Even in the case of human intelligence, there is no consensus on the feasibility of establishing its unique and uniform conceptualization [de Silveira and Lopes, 2023]. The concept of human intelligence is so difficult to define for the reason that almost everyone believes in their good understanding of it and whatever seems obvious is almost always highly non-trivial. It is amplified by the Dunning-Kruger effect (actually recognized already by René Descartes who famously and sarcastically prized God for giving everyone a sufficient amount of reason to make them happy). David Dunning and Justin Kruger empirically confirmed the correlation between the lack of competence or intelligence and the conviction of their possession [Kruger and Dunning, 1999]. Because of this effect people develop their views of intelligence consistent with their self-image, with a prominent example of a certain "stable genius".

However, the bias of idiosyncratic views on intelligence is not the main obstacle to its conceptualization. Even more confusing is the projection of common sense criteria used to assess intelligence derived from the practice (or malpractice) of assessing candidates' suitability for some tasks (including the assessment of children or youngsters applying to schools). An example of a criterion in such an assessment is frequently proficiency in "problem-solving". Not only this expression is meaningless for describing intelligence and the evaluation of such a skill is highly problematic, especially in a diachronic perspective, but it is also dangerous for carrying hidden ethnocultural bias.

It is meaningless because it is based on the assumption that the words "problem" and "solution" have a clear and objective meaning. Once again, the issue is in the illusion of obviousness. What does it mean "problem"? Here too, the answer is highly non-trivial if we want to have it comprehensive. We can try to formulate it as a call for action which can be answering a question, performing some action leading to the desired state of affairs, or some desired behavior including inaction. These are only a few possibilities out of many. To answer the question of what constitutes a solution or correct solution is an even more difficult task. The history of science gives an extensive gallery of solutions ignored or attacked by all contemporaries. Even in mathematics, the most important moments in its development were associated with changes in understanding what constitutes a correct solution to a problem (usually correct proof of a theorem). For instance, one of the main motivations for the development of set theory was justification for proofs using mathematical induction and non-constructive methods. Moreover, very often in hindsight, the solutions that violate the standards of evaluation are later or in different contexts highly prized as "thinking outside of the box" and considered as indications of exceptionally high intelligence.

There is a good example of ambivalence in the evaluation of human intelligence in the story (possibly legendary) of the solution to the long-standing problem of untying the Gordian knot that on the order of King Midas of Lycia was tied to hold the oxcart of his father attached to the column in the temple. The most popular modern story presents the highly intelligent achievement of the solution by Alexander the Great by cutting the knot with his sword. The more reliable ancient sources (e.g. writings of Plutarch and Arrian of Nicomedia) report that Alexander did not cut the knot (questionably intelligent act), but that he pulled the lynchpin from the column, uncovered the ends of the rope and easily untied the cart [Fredricksmeyer,1961]. The story in both versions (especially in the second version) identifies the high intelligence of Alexander with the elimination of complexity and the latter is clearly about the reduction of complexity of information. However, more importantly, the difference between the versions shows the difference between the ancient and modern views of intelligence.

Not only the normative idea of a good or correct solution is vague, but the development of any scale of assessment is purely conventional. With this conventionality of the assessment of proficiency in problem-solving comes the danger of cultural bias. Different cultures develop different norms and values which naturally influence the way people identify, formulate, solve problems, and evaluate these solutions. What is a legitimate problem or proper solution in one culture that is prized for being evidence of wisdom or intelligence may be considered an expression of stupidity in another.

Since all standard evaluations of human intelligence can be considered variations of problem-solving (e.g. social intelligence can be understood as the ability to solve problems in human interactions, emotional intelligence can be understood as solving problems in managing own emotions and the emotions of others) they are of little value for the general definition of human intelligence.

The same type of issue is with pragmatic views of intelligence which associate intelligence with the capacity for effectiveness in more general actions that are not necessarily in the context of problem-solving. Not only the assessment of this type can only be a posteriori, or based on the circular reasoning "capacity for effective action means having been effective in action", but also the meaning of effectiveness requires conventional, normative, and culturally loaded criteria as described above. Moreover, when we try to generalize the effectiveness of action to eliminate human aspects, we may have to accept the intelligence of objects that are unlikely intelligent. For instance, the motion of mechanical objects is effective in the sense of the Principle of the Least Action. It is difficult to accept the intelligence of falling stones because they optimize their trajectories. Mark Levi [2009] in his book "The Mathematical Mechanic: Using Physical Reasoning To Solve Problems" presents an extensive exposition of many examples of physical phenomena whose measurements can be used to find solutions to mathematical problems and of course, this does not mean that the mechanical systems involved in them solve any problems or manifest any intelligence.

Naturally, the generalization of intelligence beyond human beings is even more difficult. Alan Turing gave up this task in his inquiries of the possibility of artificial intelligence (in 1950, long before this expression was introduced, he called it machine intelligence) and proposed his imitation game (today called the Turing Test) as a functional method to judge the intelligence of artificial systems. It was based on human judgment of success in performing some tasks in which human cognitive abilities are involved, more exactly the failure of human ability to distinguish the performance of humans and machines in such tasks. In time, the test became of mainly historical interest, but the efforts to identify the common characteristics of human and machine intelligence continue the same methodological framework of comparing human and machine performance in tasks involving human cognitive functions [de Silveira and Lopes, 2023]. This performance is always related to information processing, communication, and its role in expressing behavior (execution of action). Moreover, in the case of human intelligence, the frequently invoked criterion of adaptability serves the purpose of establishing a normative characteristic associated with it. However, adaptability can be considered yet another form of the result of information communication (in the form of feedback and feed-forward loops).

Inquiries of intelligence go beyond humans or machines to include different levels of collective intelligence in individual cells, unicellular and multicellular organisms, and their populations. There is no big difference between organismal, human, and collective intelligence except for the problem of agency. In human beings, agency is associated with a conscious (rational) choice of goal/purpose and the ability to make choices between the direction of complex actions. In the absence of consciousness or the internal, centralized mechanisms of making choices, the concept of agency loses its meaning, unless we consider a mechanism at the collective level of an evolutionary feedback control reducing the multiplicity of behavioral choices of the members of the collective.

Not everyone seeks an evolutionary explanation. Michael Levin, who claims that all instances of intelligence are collective (in humans it is a collection of neurons responsible for cognition) invokes teleonomy "[...] not the final step on a continuum of agency – it is a primary capacity"[Levin,2023]. Teleonomic explanation eliminates the normative aspect of intelligence, but it does not influence the informational character of the mechanisms involved in the intelligent behavior of the collectives which can be identified in Levin's empirical study of its mechanisms. Brian Ford considers the intelligence of cells as a driver of the evolutionary process shaping the entire organisms [Ford, 2009].

Whatever the explanation of the superorganismal characteristics of collectives engaged in intelligent behavior is (using the century-old term of superorganism introduced by William Morton Wheeler [1910]), the association with an organism brings us back to human intelligence, although we don't have any more conscious or rational purpose and we don't have human physiological mechanisms. What remains are informational processes involving interaction with the environment with reduced but effective choices from a much wider range of possible states of the system.

At this point, we can enter complexity, or rather information complexity used in our definition of intelligence. The environment (in Levin's terminology problem space [Levin, 2023]) may be characterized by a high level of complex information, yet an intelligent system (individual or collective) has the capacity to reduce this complexity either in its internal modeling (e.g. in the case of human intelligence involving consciousness) or in the structures governing the behavior of a system.

### 3.2. *Information and Complexity*

The complexity of information, in turn, can be understood as a qualification of information in terms of its quantitative characteristics (the number or measure of the components of information) and qualitative characteristics (the structure of these components and the degree to which these components are bound together) [Schroeder, 2013, 2017].

Complexity must be evaluated not only in terms of the number of components but also in terms of their mutual relations. For instance, the system of preferences of an individual customer (or information about it) is much more complex than the system of preferences of a crowd of customers (or information about it), even if the latter has a much larger number of degrees of freedom and has individual customers as their components. This leads to an easy predictability of the crowd's actions in contrast to practically unpredictable individual behavior. The entire discipline of statistics is based on this distinction.

The assessment of intelligence (here, the ability to reduce the complexity of information) can be made through the observation and analysis of the overt or covert behavior of an agent. Thus, it can and should be predicated only on entities capable of transforming information. The assessment of the degree of intelligence (qualitative or quantitative) is usually possible for the agents, i.e. entities capable of goal-oriented actions, making choices, and acting based on these choices. This predication in informal contexts can be extended from the information-transforming agents to their actions or the products of these actions, but only as an abbreviated expression. So, we can say that the response to the question was very intelligent or that someone's behavior was intelligent. However, the absence of an agent disqualifies the object from being intelligent. This is why it is meaningless to say that AI is intelligent, but only that a given AI system capable of some actions executed with the use of informational interaction with its environment can be considered intelligent.

This particular choice of the understanding of intelligence and complexity will not lower the generality of my arguments as long as the reduction of information complexity in turn associated with the quantitative and structural characteristics remains relevant to intelligence. Someone who prefers a more elaborate description of intelligence may consider my definition just a terminological abbreviation. It should be noted, however, that any further specification in the description of intelligence may reduce its generality. For instance, the loss of generality by any reference to human or organismal physiology will exclude the application of this concept to collective systems of human or organismal populations, and at the same time, the intelligence of artificial systems will lose its meaning. For our purposes, it is only important to recognize that intelligence which is the central concept of AI can be identified or at least closely associated with the transformation of information and resolving its complexity.

### 3.3. *Information and Consciousness*

The relevance of information and intelligence to the study of consciousness may be obvious, but it turns out, it is highly non-trivial. Since it is a very broad topic, I will refer the reader to my detailed exposition elsewhere which here is reduced to a brief remark followed by a report of the most recent developments [Schroeder, 2011a].

The tradition of the study of consciousness as a distinctive subject of inquiry goes back at least to William James [1890]. James [1947] noted one distinctive characteristic of the phenomenal experience of consciousness – its unity. Since it was such a distinctive feature, from the very beginning it remained in the focus of all inquiries. It stimulated a direction of study seeking the mechanisms responsible for consciousness in quantum-mechanical phenomena where the superposition of states and entanglement exemplified physical processes leading to absolute unity. The vague ideas of holographic analogy became in time more specific in the search for quantum-mechanical mechanisms in the brain [Beck and Eccles, 1992].

The problems of inapplicability of the quantum formalism to the description of a large and warm brain amplified by the fact that information processing in the brain contributing to cognitive processes is distributed in its many regions stimulated attempts to modify the processes in the brain or append quantum-mechanical description culminated in works of Stuart Hameroff and Roger Penrose from the end of the 20$^{th}$ century [1996]. The attempts were not successful and although this direction of study was never completely abandoned, it is rather dormant at present.

Another direction of the study seeking an explanation of consciousness in terms of the integration of information was initiated and widely promoted by Giulio Tononi. Tononi proposed a measure $\Phi$ of consciousness understood as integrated information not only in the human brain but in any system including individual elementary particles. In the earliest papers, he did not refer explicitly in presenting his measure to integration of information, but rather to functional integration of the brain or brain complexity [Tononi *et al.*, 1994]. However soon later he made his measure a quantitative description of consciousness and the measure $\Phi$ became the central concept of what he called the Integration of Information Theory of Consciousness (IIT) [Tononi and Edelman, 1998; Tononi, 2007]. The function $\Phi$ was derived from purely statistical analysis of the simultaneous firing of neurons without any attempt to provide a structural analysis of consciousness or information integration. The mysterious non-zero value of $\Phi$ for objects such as elementary particles instead of being used as the evidence for the error in interpreting $\Phi$ as a measure of information integration became the argument for the attribution of consciousness to everything.

The panpsychism of IIT is its least problematic feature. After all, in the history of science there were many instances of contributions that contradicted common sense but later became commonly accepted. Much more serious deficiencies were in repetitions of old methodological errors and misinterpretation of the mathematical concepts used for justification of the claims. The belief that it is enough to define a measure of something to give it an ontological status is quite an extreme instance of philosophical poverty. The claim that this measure can be associated with the integration of anything and in particular with information integration, without any explanation of the meaning of the concepts of information or its integration, is another example of methodological poverty. The belief in the applicability of all mathematical concepts that involve in their name the word information to inquiries of information in all possible contexts is surprisingly naive.

More specific issues in IIT are in the complete negligence of the relationship between the external observations of simultaneity of nerve firings with the phenomenal experience of unity. The inquiries of consciousness were haunted by the homunculus fallacy for centuries. In my critical appraisal of IIT [Schroeder, 2011a], I objected to the fallacy that I called "homunculus' watch" involved in the claim that what is simultaneous in the brain for the external observer becomes phenomenal, spatiotemporal unity.

Even these clear deficiencies of IIT are not the main disqualifying features of Tononi's approach. His derivation of the measure is based on the consideration of bi-partitions of the brain for which he calculates mutual entropy which he calls mutual information. The formula proceeds to the consideration of all bi-partitions (partitions into two complementary subsets). This is a curious framework as if integration was a result of only binary interactions between the regions of the brain, while it is well known that multiple regions of the brain are involved together in every cognitive process. The binary framework contradicts the very idea of integration and makes the entire IIT irrelevant to the study of consciousness.

Someone could respond that this can be avoided by considering not bi-partitions, but multiple-component partitions representing known funcionally distinct regions of the brain. This may seem like a good resolution of the issue (never considered in IIT), but it turns out that this can produce the negative values of multiple-regional mutual information. This is an elementary information theory theorem that the binary partition is the exceptional case of nonnegative mutual information as for all higher-level partitions, even into three regions, their mutual information can be negative [Reza, 1994]. Panpsychism is a bizarre, but still conceivable view of reality, but probably nobody would accept negative consciousness.

The IIT was promoted with so noisy fanfare that its proponents and supporters did not hear the criticism "of the leading theory of consciousness" growing over the years until the very recent burst of the bubble when 124 researchers signed the open letter calling IIT a pseudoscience [Fleming et al., 2023]. This is an extremely embarrassing and unprecedented situation in science. In response to this letter, several leading researchers of consciousness who in the last quarter of the century never voiced any objections to IIT suddenly worry that the use of the word pseudoscience is "unfair" for what they admit now actually is "bonkers" and that it may slow down research in this important subject [Lenharo, 2023].

This type of defense is as bizarre as IIT. The elimination of erroneous methodology will not slow down but rather accelerate progress. The criticism is of the errors made in a particular approach of particular individuals and of the uncritical acceptance of its products, not of the inquiry of information integration. The harm was done not by the critique of the errors, but by the lack of it in the quarter of a century time when IIT was promoted as a "leading theory of consciousness". The more than a century of inquiries of consciousness as a process in which information is integrated will not be wasted when the alternative already existing and future approaches correct old errors.

The key lesson from this dramatic development is that the study of consciousness as integrated information requires a good understanding of what information is, what its integration is, expressed in the proper theoretical description, and proper formalization [Schroeder, 2009].

## 4. Conceptual Obstacles in the Study of Information

### 4.1. *Defining Information*

The association of intelligence and information is mutual. Not only we can study intelligence by making inquiries of the way information is transformed and used, but the other way around, we can consider an intelligent way to perform inquiries of information. The attempts to define information are simply instances of more or less intelligent actions to organize a very broad range of phenomena with similar characteristics into a unified conceptual system with lower complexity.

These attempts have to satisfy two conditions. The first condition is that the concept of information has to be sufficiently general and at the same time sufficiently specific. It has to include all unquestionable instances of the use of the term information. Any definition of information that does not apply to language and other forms of communication, semiotics, processes of genetic inheritance, processing of information in living organisms and their populations, mechanisms of control and governance, computing devices, and other contexts of the use of the term information cannot be considered adequate in the general study of information. On the other hand, it is necessary to distinguish between the concept of information and concepts such as knowledge, belief, opinion, wisdom, etc. Thus, we have to avoid over-generalization.

The preceding sections of this work provided arguments for the study of information as a fundamental tool for inquiries of intelligence, complexity, and consciousness. Such inquiries require a very high level of generality exceeding specific interests in intelligence in the context of life not only because we want to consider intelligent information systems designed and implemented by humans as a part of the technological progress that we cannot predict, but also because we have to have intellectual tools for the search for extraterrestrial intelligence. There is no reason to claim that intelligence requires the forms of life that we know from our direct environment. We have to consider the possibility that life can have very different forms, based for instance on an alternative configuration of chemical elements. Moreover, we even cannot claim that intelligence requires its implementation in life forms.

There is an additional condition for the definition of information, much less obvious and frequently challenged that it should follow the rules of logic. The form of inquiry of information is a matter of choice, i.e. convention, and nobody is prevented from other forms of inquiry, including artistic expression of subjective, intuitive perceptions of informational phenomena. The choice of logical rules for defining concepts is simply a matter of the style of inquiry. This may be considered unnecessarily restrictive. After all, the famous Bateson's "definition" of information as "difference that makes a difference" by the use of an idiomatic expression "makes a difference" is metaphoric and far from being logically and ontologically correct (once again, its strictly logical interpretation leads to a category error).

Bateson intentionally made it open-ended to extend its generality and without a doubt achieved great success in finding followers who accept it as a definition while presenting very different and frequently contradictory interpretations. Batson himself presented at least half a dozen different interpretations with increased logical precision, but each of them significantly restricted the generality of the concept of information. However, it is possible to choose some reformulations of Bateson's information to make it a well-defined concept [Schroeder, 2019]. Therefore, with admittedly lower precision Bateson's information is sufficiently close to meeting logical requirements.

A much bigger problem is with other attempts that seemingly follow logical rules of defining concepts, but which use as defining concepts equally unclear undefined concepts or concepts that in the language of the discourse are species of the genus information (e.g. "data" which means "given" in Latin, and for which it would be difficult to avoid invoking full expression "given information"). The requirement to avoid the use of undefined concepts in definitions is obvious, but not sufficiently obvious to prevent the import of apparently obvious common-sense words or expressions that everyone can interpret freely (such as "difference").

If we follow the directives of the formulation of logically correct definitions avoiding under- and over-generalization of the concept of information, the large variety of attempts is reduced to only a few that did not gain much popularity. It should not surprise us that the discussion of the concept of information has never been finalized by those engaged in this inquiry and the majority of people using the term information are not even aware of the absence of a commonly accepted definition.

Finally, it has to be stressed that there is no need or justification for the claim that there is only one "correct" definition of information. Definitions can be logically correct or incorrect, but only theories that are built over those definitions can be tested and evaluated. There is no nontrivial concept in the history of science or philosophy with an uncontested, single definition. Thus, the choice of definition matters only when it is a part of developing a theory of the concept.

### 4.2. *Misunderstanding of the Concept of Information*

Claude Shannon did not define information or even make a distinction between information and uncertainty and could develop such a successful theory of communication that his followers convinced him to rename it "the" theory of information, so why anyone should care? [Shannon, 1948; Shannon and Weaver, 1949/1998] After all, Shannon provided three principles for the quantities H measuring how much choice is involved in the production of discrete information in the form of a sequence of events with some probabilities. The principles determine the functional form of H in terms of the probability that "play a central role in information theory as measures of information, choice and uncertainty" [Shannon and Weaver, 1949/1998] and which are similar to entropy in statistical mechanics [Shannon and Weaver, 1949/1998]. What Shannon called "entropy of the set of probabilities" [Shannon and Weaver, 1949/1998] became a powerful tool in the study of communication with a myriad of applications. Since it gives a quantitative description of something, why not dispose of the choice and uncertainty and settle on this something being information?

Not everyone agreed with this idea and very soon the price for the hidden assumption in Shannon's principles that the order of events in the process of production of information is irrelevant became the source of criticism [Bar-Hillel and Carnap, 1952]. Is the measure of information in the words "dog" and "god" and in a meaningless sequence of letters "ogd" really equal? Shannon prevented such criticism by declaring that the semantic aspects of communication are irrelevant to the engineering problem and that therefore the words dog and god differ only in their meaning. It is easy to agree that the semantic aspects of communication are irrelevant to the engineering problem of the speed of transmission of information in communication. However, it is clearly false when other engineering problems are considered. Can we consider an engineering problem of efficient transmission of information solved by sending a report about the number of occurrences in the message for all characters in the alphabet or their relative frequencies? The entropy will be the same as for the original message, but the message, if long would be completely lost in transmission.

Shannon was aware of the issue even if he did not write explicitly about it in a critical way. He wrote in his famous paper entire two sections, Section 2 "The Discrete Source of Information" and Section 3 "The Series of Approximations to English" about the analysis of the sequences forming messages using conditional probabilities of the choice of a letter or word based on preceding letters or words [Shannon and Weaver, 1949/1998]. In 1948 the task was too difficult to have any practical application for the structural analysis of information. It may be surprising that in some sense the generative AI systems can be considered a form of realization of Shannon's idea with the training of neural networks replacing the calculation of conditional probabilities.

The calculation of conditional probabilities even now would be unrealistic. Instead, generative AI systems such as ChatGPT are trained with the methods of deep learning on the large data set from the internet to acquire the ability to choose which character or word should be selected next in the generation of text. This is a much better solution to an engineering problem to generate meaningful responses to inquiries, but we are not closer to the methodology of structural information. The process of training does not involve any structural analysis of information or its semantics as it functions as a "black box" without accessible memory. It is just reproducing typical (i.e. highly probable) structures of the instances of information in the training reservoir. The fact that the choice is highly probable does not mean that the sentences in the generated text are true or that they make any sense. These hallucinations of generative AI systems demonstrate that Shannon's idea of replacing the structural analysis of information with probabilistic methods was faulty.

Shannon was aware of the importance of the order of characters in the message (i.e. the structure of a message) which he believed could be described by conditional probabilities. However, he did not provide or even try to find any tools for the structural analysis of information. One of his principles for entropy was that the order of characters does not matter. Moreover, he believed that his main achievement was going beyond what Ralph Hartley did ten years earlier by assuming that the characters have equal probabilities [Shannon and Weaver, 1949/1998]. Did Shannon read Hartley's article? He cited it, but the word "probability" does not appear in Hartley's article [1928] and this concept does not play any role in it which is in clear contradiction to Shannon's interpretation that Hartley assumed equal probabilities.

Contrary to what Shannon wrote in the introduction to his paper, Hartley [1928] did not use the assumption of equal probability of characters (or probability of anything else) to derive the logarithmic measure of information. He simply observed that the encoding of information can be optimized, i.e. changed without any essential information change. According to Shannon's conceptual framework this optimum is achieved for the uniform probability distribution of characters. However, Hartley referred not to probability but to the experience of operators encoding messages who certainly knew about Morse's optimization of encoding based on the frequency of characters in the language.

Hartley [1928] derived his formula corresponding in probabilistic interpretation to the special case of Shannon's for uniform distribution from the assumption that the measure of information should be invariant with respect to the change of its encoding. Hartley programmatically avoided any use of psychological considerations such as the way human beings achieve an understanding of the meaning, but his view that the same information can be encoded in one or another language, or using encoding systems with different numbers of characters indicates that information is invariant with respect to such changes, although he did not refer explicitly to the meaning.

Yehoshua Bar-Hillel and Rudolf Carnap [1952] were probably the first to reject Shannon's theory as a theory of information because of its disregard for semantics. They considered Shannon's work (quite rightly) a study of signal transmission and attempted to develop a theory of information based on the logic of language. Their proposal of the logical theory of information equipped with semantics was not very successful in directing further research.

My diagnostic [2012] of the limited resonance of their approach in the literature on the subject of information was that despite the promises to deliver a semantic theory of information as semantics is understood in the logic of language, their approach was still syntactic. This fact was hidden in the substitution described in one sentence declaring that "for technical reasons" they replaced the states of the world addressed by information with their descriptions. Surprisingly, this shift from semantic to syntactic analysis of information was usually overlooked.

As nobody ever noticed that despite the declaration that their theory of information was semantic, it was actually syntactic, probably a more direct reason for the lack of popularity of the approach proposed by Bar-Hillel and Rudolf Carnap was in their attempt to direct the development of the theory to arrive at results comparable to those of Shannon. As a result, it was not clear how their approach was better. At least, it did not resolve the issue of the meaning of information.

The logical tools of their theory did not help much as logical semantics had more questions than answers at that time. Because they substituted syntax for semantics in their considerations without excluding intentionality as a basis for meaning, they did not identify the actual source of difficulties and did not overcome these difficulties in the search for the meaning of information. At least, they were the first who, using

Shannon's expression, did not "jump the bandwagon" and openly criticized his programmatic disinterest in the meaning of information.

The prolonged discussion of the conceptualization of information and the lack of consensus on its definition is not a problem. It is just evidence of its importance and relevance. The much bigger problem is that the diverse attempts to formulate a definition are rarely followed by developments of comprehensive theories of information. Competing definitions of information can be evaluated only through comparisons of the theories of information based on these definitions. Regretfully, the definitions of non-trivial concepts are rarely compared using the criterion of the explanatory power of the theories in which they were used.

We could see that the main unresolved (or not satisfactorily resolved) problems in the study of information were related to two somewhat related obstacles: the lack of tools for the structural analysis of information and insufficiently general semantics of information that usually mimicked linguistic semantics. The minimal criterion for an adequate conceptualization of information is to allow the development of a theory of information that helps to overcome these obstacles.

## 5. Semantics of Information

### 5.1. *Attempts to Develop Semantics of Information*

After the initial attempt by Bar-Hillel and Carnap to develop a semantic theory of information which already was limited to information in its linguistic form in which logic could be employed, there were not many other attempts to develop a general semantics of information not limited to particular contexts. This should not be a surprise considering the formidable task of answering the question about the meaning of meaning in the limited context of language and human comprehension that remained not achieved in several centuries despite multiple attempts.

One of the reasons for the difficulties in understanding meaning came with its association with intention, a mysterious capacity of the mind to cross the Cartesian precipice separating *res cogitans* and *res extensa*, the mental and material realms, reaching from the thought in the former realm to point at the denotation residing in the latter realm. Cartesian duality was the main obstacle to understanding consciousness which resided in the former realm but was influenced by the objects from the latter realm and in turn could affect these objects. In the case of intention, the situation was even more complicated. Signs belong to the realm of *res extensa*, but they acquire their symbolic characteristic only after they are interpreted by a conscious subject in the realm of *res cogitans* pointing at their meaning in the objects among *res extensa*. The divide between the two realms had to be crossed twice. This led to the most typical tripartite models of semiosis with the *sign*, *interpretant*, and *denotation* (the last one acquiring the status of the meaning of the sign) in the terminology of Charles Sanders Pierce's *semiotics*. For Peirce [2015], it was a tertiary relation expressing a cooperative action not reducible to its binary components. In his explanations, interpretant was called sometimes interpretant sign as it was rather the effect of the sign on some agent (quasi-mind), not necessarily mind and itself can serve as a sign.

Peirce [1977] explicitly wanted to maintain a higher level of generality by not limiting the interpretant to a conscious person, using this simplified interpretation of the interpretant to make his explanation easier. However, the generalization could require a sequence of tripartite relations compounding interpretant signs over interpretant signs considered as signs. The tripartite relation is more of a fundamental framework underlying the process of semiosis. On the other hand, the consistent use of the expression of the semiotic process places semiosis within time and space limiting its generality.

This attempt to maintain a level of generality exceeding the psychological explanation of the intention as a mental phenomenon was unusual at the time. Franz Brentano [1874/1995] who followed the Cartesian cut made a clear distinction between what he called, using the Scholastic terminology, the intentional or mental inexistence of an object directing towards an object which was an exclusive characteristic of mental phenomena always including something as an object in themselves, as opposed to "physical phenomena." Obviously, the reference to mental phenomena reduces generality to the purely psychological level.

The subject of "The Meaning of Meaning" [1923] was brought to the attention of a broad intellectual audience by the book of this title published by Charles Kay Ogden and Ivor Armstrong Richards. Their approach was again based on the triangular scheme involving symbol, thought, and referent engaged in the binary relations thought – symbol qualified by correctness, thought – referent qualified by adequacy, and symbol – reference qualified by truth. The triangular scheme described the instrument of both human communication and thought in a culturally determined context. Although their approach introduced yet another aspect of culture to the study of meaning, it did not lift the generality of the perspective above the level of the use of language by human beings.

The first substantial generalization came with the very rich direction of the study in which "*bio* joined *semio*" (using Kalevi Kell's expression for the early biosemiotics [1999]) initiated by Friedrich Salomon Rothschild [1962]. Rothschild [1962] did not even mention information in his three laws of biosemiotics investigating "the communication processes of life that convey meaning in analogy to language", but his work was already preceded by Erwin Schrödinger's epoch-making small book "What is Life?" [1944] that stimulated the revolution of genetics so that it was just a matter of time for biosemiotics to become information theory for living organisms in their complex multilevel architecture from molecular to organismal and beyond to their populations.

In the immensely extensive literature on biosemiotics, there is a very frequent reference to the meaning of information, but always within the context of its function in life at some level of its organization. The meaning of information becomes a secondary concept explained by its function in either a causal, deterministic, or teleonomic way. Although the tripartite semantics of the earlier authors is not necessarily invoked directly, it is hidden in the ecological framework. Life at any level of organization cannot be considered without its environment and the constraints imposed by it. An additional limitation of the biosemiotic information at the organismal level comes with agency characterizing living objects. Without ecology and agency, the meaning of bio-information is losing its meaning. This of course does not make biosemiotics inferior or incomplete, but it shows that in looking for context-independent semantics free from the triangular relation engaging pragmatic aspects we cannot simply import the bio-semiotic framework.

There is nothing wrong with crossing the border separating semantics with pragmatics. The commonly invoked borders between syntactic, semantics, and pragmatics in the study of language popularized by Charles Morris in his useful classification of the subjects in the study of language may lose their application in the case of information [Posner, 1992; Schroeder, 2011b]. Even in the study of language, the use of language (i.e. the subject of pragmatics) dominates the inquiry of the meaning (semantics). A prominent example is in the "language games" of Ludwig Wittgenstein [1953].

Thus, the problem of using pragmatic explanations of meaning in the context of information (which includes language as only one of the possible information systems) is not in blurring the divisions between the domains of linguistic studies, but in the relativization of the meaning to the user of information. While someone could insist that every language requires a user or rather a community of users, in the general study of information this assumption brings an unacceptable restriction of the concept of information. If we want to consider information understood in its sufficient generality explained above in this article, applicable to information systems in the early stages of the universe before the existence of any forms of life became possible or in the regions of the universe where even now this existence is not possible, we have to eliminate the tripartite framework of semantics engaging concepts of a user, interpreter or thought.

Some attempts to present the semantics of information or the semantical concept of information referred to the idea of "the true information". If we want to include in the theory of information language as a special instance of information the attempts to define the truth are doomed by Tarski's Theorem on Undefinability of the Truth [Tarski, 1983]. The statements about the truth can belong to the metalanguage but cannot be expressed within the language of theory. Thus all attempts to define information with the use of the qualification of being true, for instance,

"true data", do not make much sense unless again we restrict our inquiry to a particular context in which the linguistic form of information is excluded giving us freedom from the constraints of logic.

This is conflated with another issue arising in using data in the definiendum for information as if information could be considered a qualified type of data against the etymology of this word and virtually all its applications. Can we consider data that are not information? If not, then data and information become synonymous. If yes, what are they, and in what sense they are given? It can be claimed that data that are not information are false. Then we are trapped in the undefinability of truth.

In the common-sense view, "true" can be interpreted for instance as "effective" in achieving some goals. However, in this situation, we can simply eliminate the former as it just creates confusion by the suggestion of unjustified generality without adding any explanatory power. We can find an analogy with the infamous explanation by Herbert Spencer of the Darwinian concept of natural selection as "the survival of the fittest". Which species are the fittest? Those are that survived. Which instances of the information are effective? Those that are true. Which are true? Those that are effective.

Thus, there is no sound and sufficiently general semantical theory of information. I presented the possibility of overcoming the difficulties in my earlier publications [2011b]. The proposal was based on the assumption that the obstacles in the development of the theory of intention (aboutness) were created by the faulty view that elements of the language (e.g. nouns) are about objects that are entities of a very different ontological status. The correspondence between symbols and their denotations requires a theory overarching the study of both types of entities which seems impossible.

There are two possible solutions to this obstacle. We could consider uninterpreted signs and their denotations as entities of the same type (physically created with the ink inscription "dog" on a paper and an instance of an animal in our environment that is a dog) and then the correspondence is between entities of the same status. This could be one of the possible interpretations of the substitution made by Bar-Hillel and Carnap when instead of considering the states of the universe they involved the descriptions of such states.

In this case, we have to consider the description as an object of the same ontological status as the described object. If we claim that the object and its description are both devoid of informational content and that information has a relational character of secondary ontological status, then we encounter the old obstacle of mysterious correspondence between a sign and its meaning. This correspondence is purely arbitrary (there is nothing in the reality of these two types of entities that connect them) and it is not one-to-one. Since we can have multiple sign systems addressing the same entities and signs may have some level of abstraction addressing for instance many animals that we consider dogs, the theory of intention within the reality of denotation breaks up.

There is another way to overcome the difficulties of crossing the precipice between entities of different ontological statuses. This solution is possible at the level of the study of information, but not at the level of language. Intention can be considered a relationship between two informational systems. One of them is symbolic, the other possibly not. Thus, the word "dog" is about, not an entity of a different ontological status "dog in itself" (paraphrasing Kant's "thing in itself"), but about the information associated with this "dog in itself". We can only claim that the object of our inquiry is a dog (we give meaning to the object) through information carried by this dog (for instance, how it looks, how it smells, what sounds it makes, etc.) which is transmitted to us and perceived by us. These are our cognitive mechanisms that link the perceptions of the information about the writing "dog" on the paper and the perceptions about the animal. On both sides, we have only informational entities. Their correspondence is established based on their structural characteristics. We cannot apply this to the linguistic logical theory because we cannot claim that an animal that we call a dog is a linguistic phenomenon. After all, dogs existed before languages developed. However, nothing prevents us from saying that the inscription "dog" stands for the information carried by some type of animal or an instance of such an animal.

### 5.2. *Reverse Semantics and Encoding of Information*

The second solution of interpreting meaning as the relationship between informational entities can be understood as "reverse semantics". The adjective "reverse" refers to the change of the paradigm in linguistic semantics which starts from the pre-existing language (of any type) with fixed rules and vocabulary and proceeds to establish a way (intention) in which entities and their relations are represented within language.

This is in my view placing a carriage ahead of the horse. In the approach proposed here, the starting point is in the inquiry of the informational structures of entities and the role of semantics is to inquire a variety of ways in which this information can be encoded. Some of these encodings may have the form of a language, but encoding should not be considered a human or intentional action. The meaning of information in its linguistic form consists not of the entities of reality devoid of information content, but of informational structures which are independent of the language.

The study of these informational structures may involve physics but is not limited to the objects of study in this domain. Reality has a hierarchical architecture of multiple levels of complexity irreducible to simpler ones. The study of these levels (for instance in living organisms) can be supported by physics but cannot be reduced to it. Yet, in one respect physics can be a resource for the study of other levels of complexity due to a long experience in the inquiry of symmetry. After all, the invariance of information with respect to the transformation (change) of encoding is a form of symmetry and symmetry is one of the most important tools of science with a long tradition of its use.

This idea of intention as an informational relationship is not completely detached from the earlier ideas expressed in terms of linguistic logic but of course, these other older ideas were expressed without the use of the term information or concept of information.

First, let's notice that the trick employed by Bar-Hillel and Carnap in their attempt to formulate a semantic information theory was to some extent similar. They replaced "the state of the world" with "the description of the state of the world" [Bar-Hillel and Carnap, 1952]. This way, in their approach, although they considered this a marginal, technical procedure, information was not about the state of the world but about the description of the state of the world, obviously an informational entity. The problem here is that in their approach the description of the state of the world requires already existing semantic correspondence which they did not elaborate on as just a secondary technical issue. However, this way, when they assume the existence of the description, their reasoning becomes circular.

Much closer to my approach presented above was John Stuart Mill's view of the meaning [1843]. Mill involved in his description of the meaning of a term two concepts, of its denotation (basically similar to what commonly is understood by meaning, i.e. the set of entities to which the term can be predicated) and connotation (the organized system of predicates or properties which can be predicated on all entities from the denotation).

We can identify a similar idea in the works of Peirce [1867], although his view was expressed in a metaphorical way of multiplication of numbers, that information is equal to the product of extension and intention ("breadth $\times$ depth of the concept"). Although Peirce considered information in a rather narrow context of the characteristics of concepts, it is interesting probably the first occurrence of this term in the relevant literature.

These views are surprisingly intuitive and at the same time surprisingly absent in the contemporary discourse of meaning. In particular, Mill's view that it is the connotation that determines the meaning is of special importance. We cannot comprehend the denotation which may consist of a large or even infinite number of entities and the only way we can understand the meaning of the term is by the connotation that gives us characteristics of the objects in the denotation. We know what the meaning of the word "dog" is not because we know all dogs, but because we know what properties of dogs make them dogs. The connotation can be easily identified with the informational content of the term.

Moreover, the Aristotelean genus-species (classical) concept of definition is based on a similar but slightly modified approach. Aristotle refers to the connotation of the defined concept (genus) and then seeks differentia, a description of differences that distinguish the defined concept from all other species of the genus. This approach refers to the same order structure but involves an additional instrument – differentia analogous to Bateson's difference. His "that makes a difference" can be interpreted as the distinction of differences of species within the genus. Which differences do not make a difference? Those which involve species not included in the genus.

We have here the first Bachelardian epistemological rupture tearing down the separation of semantics and syntactic of information. The next step in removing epistemological obstacles is the reversal of the relationships. Orthodox semantics starts from the pre-existing language and assigns intention to its components (terms) that carry meaning. This is against the historical order of affairs. At first, there were objects of our (human) comprehension whom we gave informational nature as carriers of information. Then gradually humans associated with them (also informational) symbolic entities. This we can call encoding information associated with the process of assigning symbols (thus, reverse semantics), but the original information was already encoded (without our awareness) in the objects of our comprehension. These objects are our construction in the sense that our comprehension selects the information characterizing them, i.e. information encoded in them. Thus, instead of focusing on intention as action directed to the entities of external reality, we should focus on encodings of information in different information systems (consciousness, entities, etc.) and the correspondence between structures of these encodings. This approach emphasizes the structure of information which can be analyzed through its manifestation in encoding. The isomorphism of the structures defines meaning.

In this perspective, Hartley's observation of the invariance of information with respect to the change of encoding acquires special importance [Hartley, 1928]. Although he considered this invariance as obvious and wrote about it mainly for the derivation of his measure of information, the recognition of the variety of encodings of the same information makes his study the actual initiation of the study of information. Of course, it would be an anachronism to claim that Hartley was aware of the fundamental issues of the semantics of information and that he appreciated the role of invariance of information in its different encodings for the development of the study of information. For him, it was an obvious characteristic of information derived from the practical experience of telegraphy useful for the measuring of information. However, the way he approached the matters of encoding information in the contexts not only of texts but also images shows that his approach based on an intuitive insight formulated twenty years before the publication of Shannon's work in some aspects was superior.

## 6. Qualitative and Quantitative Methodologies

One of the most common myths about science is the belief that numbers are the ultimate tools of scientific inquiry and that quantitative methodologies of inquiry are always better than qualitative ones. I already published a critique of this unfortunately very popular view of science and will not repeat its arguments here [2020b]. I will focus here on the methodologies of the study of information. Certainly, the popularity of this view among the members of the general audience is perpetuated by the false opinion that mathematics is a discipline studying numbers. Media, with the best intention, promote the idea of "numeracy" as an important part of education or lament its decline in contemporary societies. Teachers are warned about the disability called "dyscalculia" as a source of difficulties in learning mathematics.

Of course, there is a very important sub-discipline of mathematics called number theory with multiple applications across many other sub-disciplines. However, there are many other sub-disciplines where number theory is absent or does not play any important role. Moreover, number theory studies not numbers but structures defined on sets of numbers. It is powerless to answer the questions about particular numbers although "numerologists" (not "number theorists") will tell you a lot about terrible numbers such as 4, 13, and 666 (the last is the worst for sure). I am sure that all promoters of numeracy would be terrified if they learned that the arithmetic of natural numbers or the theory of real numbers cannot answer the question about the result of 2+2 without a prior clarification of the convention in writing names of numbers. The result can be 4 (if we use the decimal convention) or 11 (if we use the convention with three digits inherited from the decimal system). Number theory does not deal with the questions about the result of 2+2, but about what we can say about numbers when we eliminate conventions.

The enthusiasts of quantitative methodologies frequently refer to Galileo Galilei's view on the use of mathematics in the study of nature: "In *The Assayer*, [Galileo] wrote 'Philosophy is written in this grand book, the universe ... It is written in the language of mathematics, and its characters are triangles, circles, and other geometric figures;...'" [Drake, 1957]. There is nothing about numbers here. Of course, Galileo was a pioneer in experimental methodology and his work involved measurements and therefore numbers. But his understanding of numbers was still in the tradition making them derivative from the geometric intuition. It took an additional three centuries to make real numbers independent from geometry, mainly through the work of Dedekind who provided the description of their construction now known as "Dedekind cuts" in the second half of the 19$^{th}$ century.

It was the time when the infatuation of scientists and philosophers with numbers became common. A prominent example of this infatuation can be found in William Thomson's (Lord Kelvin's) *Popular Lectures and Addresses (1891-1894)*, where he wrote: "When you can measure what you are speaking about, and express it in numbers, you know something about it; but when you cannot measure it, when you cannot express it in numbers, your knowledge is of a meager and unsatisfactory kind [...]" [Thomson, 1981].

Although the fallacy of the belief that mathematics is about numbers and not much else is so common, it is not more dangerous than any other expression of ignorance. Much more dangerous are false beliefs about some concepts in mathematics held by people with a relatively high level of familiarity with mathematics, frequently even teaching mathematics at some levels of education.

A typical example of such false belief continuously promoted in the majority of introductory college textbooks for algebra is that the set of real numbers has an easy explanation by its one-to-one correspondence to the set of points on a straight line. We have to choose one point representing 0, and another different point representing 1. The increase in numbers will be in the direction of 1 from 0. Then we relate with a non-negative number x the point on the side of 1 which is in distance equal to x. If x is negative, we relate to it the point in the distance equal to the absolute value of x on the left side from 0. Conversely, if we have a point on the side of 1 we relate to it the real number equal to its distance from 0. For points on the other side, we relate to them numbers opposite to their distances. So, it is concluded that we showed that points of the straight line and real numbers are in a one-to-one correspondence and the order of numbers corresponds to the order of points.

In more than forty years of teaching math to undergraduate students (some of them very intelligent), I never received a correct answer to the question of why this reasoning is completely invalid and without any hope of making it valid by revisions. It was not the students' fault. In addition to being confused by the deceptively intuitive terminology that can fool even more experienced learners, they were already brainwashed in their high schools. It did not help that before the class about this so-called "real line," I always warned the students about the error of circular reasoning.

Students are equally surprised when I demonstrate that the geometric constructions in the style of Descartes cannot help in the determination of the position of points corresponding to all but a small, countable subset of real numbers. It does not help to consider all computable real numbers or numbers that can be identified using any logically formulated description (again both are countable subsets). Any reasoning about all real numbers involving as one of the premises "If we know what x is, then ..." has to be invalid, because the majority of real numbers cannot be identified. The theory of real numbers is about the structure of real numbers (formally called the field of real numbers) concerning algebraic operations, not about individual numbers. Thus, the "real line" does not prove anything and does not explain much about real numbers. It only explains how we can create a consistent, but rather arbitrary model of the geometric one-dimensional space using an algebraic conceptual framework when we assume that there is some measure of distance defined on it. Of course, without assuming a specific measure there is no correspondence between real numbers and points of the line.

This demonstrates that the belief in the superiority of the quantitative methodologies involves the fallacy that the association of the elements of reality with numbers has an explanatory power because we can comprehend numbers easily, although this comprehension may be illusionary. However, there is another fallacy involved in the opposition between the quantitative and qualitative methodologies. It is based on the conviction of their essential difference. In reality, the mathematical concepts associated with numbers such as the concept of a measure or a distance are simply tools for modeling or generating structures within the field of scientific inquiry. Specific numbers do not have any meaning, only their mutual relations invariant with respect to some transformations.

Numbers obtained in measurements involve conventions of the choice of the system of units. The measurements serve the purpose of determining the structures of investigated objects and phenomena in which these objects are engaged. The structures are not numbers, magnitudes, or quantities and therefore structural analysis belongs to qualitative methodologies. Quantitative methodologies are just components of more general qualitative methodologies. The confusion about the distinctive forms of inquiry arose from the fact that some forms of inquiry, especially in the contexts of the lower level of abstraction, may involve very simple methods of the collection and analysis of information that do not require advanced mathematical formalisms. Typically, the low level of abstraction is associated with structures that do not require numerical tools. However, there are also many forms of structural analysis engaging sophisticated mathematical theories in which concepts of numbers are absent. There is no reason to claim the essential difference between quantitative and qualitative methodologies and no reason to claim the superiority of either of them.

## 7. Digital and Analog Information

The most popular distinction between digital and analog computing introduced by von Neumann [1963] is based on the difference between the symbolic, digital, and discrete representation of numbers and their apparently continuous representation as magnitudes characterizing the states of physical objects. In reality, although this distinction is highly intuitive and seems to reflect objective differences between the forms of information, it is purely conventional. The functioning of all computing devices involves the manipulation of the physical states of their operating systems and at the same time, the digital representation of numbers is achieved by a conventional discretization of continuous magnitudes.

In this paper, I am using the distinction between analog and digital information and computing that I introduced in my earlier work in which the difference between analog and digital information is similar to the difference between the concepts of physics characterizing physical systems by physical states (analog) and observables (digital). This distinction in physics acquired fundamental importance with the rise of quantum mechanics but was already present earlier. I wrote "similar" because essentially identical distinctions can be identified elsewhere. For instance, the foundations of probability theory can be built starting from the concept of a family of events understood as measurable subsets of an outcome space and proceeding to random variables, alternatively, starting from an appropriate algebraic structure of random variables and proceeding to special class of random variables that can be interpreted as characteristic functions for subsets of an outcome set corresponding to events.

The depth of the distinctions between the fundamental concepts that we can interpret as a state of the system and that of observable in both quantum theory and probability theory is rarely recognized, at least not in an open way. In quantum theory, the issues are hidden for instance by the use of ad hoc terminology of "hidden variables" (that sounds better than the oxymoron "unobservable observables"). More recently, to avoid the name "observable" suggesting engagement of human inquirer, the name "beable" was introduced by John Bell. In probability theory, the standard trick to avoid complications is to focus on just two special cases of discrete and continuous random variables while excluding anything else that causes trouble. Probably the closest to an honest denunciation of the forgotten problems was the series of the 1998 Turin Lectures by Gian-Carlo Rota [2001] *Twelve Problems in Probability No One Likes to Bring Up*. A more extensive and detailed exposition of some issues addressed by Rota is in the book created in collaboration with others [Kung *et al.*, 2009].

Rota's lectures addressed not only issues within probability theory but also the study of information. In particular, Rota addressed the issue of the formulation of the orthodox information theory derived from probability theory while it should precede probability. This concern was not new as it was already voiced by Andrey Kolmogorov [1983] a long time ago when he proposed his solution in the form of a description of algorithmic complexity. Kolmogorov's approach did not bring a sufficiently general solution and Rota directed the future inquiry toward a new logic of information in terms of the lattice of partitions.

A similar issue is in the relationship between quantum physics and quantum information theory. Quantum computing became the hottest topic of this century but it seems that here too the carriage was placed in front of the horse. The usual approach is to study quantum computers considered as a special case of a quantum system and quantum information is just an engineering concept necessary for the use of such computing devices. With the increasing role of information as the most fundamental physical concept as promoted by Rolf Landauer [1991, 1996, 1999a, 1999b] (*Information is Physical*) and John Archibald Wheeler [1990] (*It From Bit*), quantum theory should be derivable from quantum information theory [Wheeler and Ford, 1998].

The main problem is in the main focus on quantum computing formulated almost exclusively in terms of qubits, i.e. quantum systems that are described in terms of two-dimensional Hilbert spaces. The description is appropriate for quantum computer systems built with processors whose state is described in terms of spin. However, the theory of qudits (quantum information systems that can serve as quantum logical gates described by Hilbert spaces of dimensions higher than two) which was initiated at the end of the 20$^{th}$ century remains in the status of future or early inquiry [Rains, 1999; Gottesman, 1999; Vourdas, 2004].

The source of the problem is a well-known but frequently ignored fact that quantum theory in two-dimensional Hilbert spaces is fundamentally different from all cases when the dimension is higher than two or infinite. For instance, Hilbert spaces of dimension three or higher have to be infinite if we want to have orthocomplementation defined for their subspaces, while there are finite two-dimensional Hilbert spaces with orthocomplementation. Since obviously the restriction of the quantum theory to one special binary case does not make sense, there is no hope that quantum mechanics can be derived from the quantum information theory developed in terms of qubits.

A closer look at the formalisms involved in quantum and probability theories brings into focus another similar type of formalism developed in the semantic inquiries of modal logics based on the idea of possible worlds (initiated by Rudolf Carnap but already considered by Leibniz), in particular in the frames of Kripke semantics. Semantics in logic is introduced with the use of truth/false valuations of sentences in the two-element Boolean algebra. This can be implemented through functions on the set of all sentences with values in a set {0,1} with appropriate conditions of consistency distinguishing the sentences with values 1 as (descriptions of) possible worlds. Then the necessarily true sentences are those that have valuations 1 in all possible worlds, and possibly true sentences have valuations 1 in at least one possible world. Thus, a similar semantics for information in the style of Kripke semantics can be developed in terms of valuations. However, this can be done only with the prior development of the logic of information.

## 8. Brief Outline of an Example of Theory of Information Meeting the Postulates of the Rupture

### 8.1. *Definition of Information*

Thus far, this article presented a critical analysis of methodological assumptions inhibiting progress in the study of information as a fundamental concept for inquiries of intelligence, complexity, and consciousness together with postulates to eliminate obstacles. There is a legitimate question about how realistic these postulates are. This section is intended as a confirmation that a theory of information meeting the postulates is possible.

The remaining part of this paper presents an outline of a theory of information developed in my earlier publications formulated here in a way consistent with the postulates promoted in this work [2011c, 2022]. The reason for this short presentation (extensive and detailed presentations are published elsewhere in my multiple articles) is not intended as a closure of the theoretical study of information, but rather its opening at a sufficiently high level of generality for inquiries of intelligence, complexity, consciousness, etc. Its direct objective is to demonstrate or rather illustrate practical applications of the methodological tools described in this paper. This illustration of the use of methodological tools may be helpful for alternative conceptualizations of information.

At this point, it is important to recall that there are some alternative approaches to the study of information that explicitly denounce the shortcomings of the Shanonian tradition and offer ways of their elimination. It would have been too extensive a task to present them all in this work. Since none of them resolves or addresses all the issues analyzed here, the present analysis may be helpful in their continuation. The approach with the most interesting results in my subjective judgment, an advanced theoretical formalism, and applications in more specific domains was developed by René Thom in his famous but currently rarely and insufficiently revisited book "Structural Stability and Morphogenesis: An Outline of a General Theory of Models" [1975].

Thom provided excellent tools for the structural analysis of information. His way of thinking about the relationship between qualitative and quantitative methodologies was similar to that presented in this work for instance in his criticism of Rutherford's dictum "Qualitative is nothing but poor quantitative."

The content of this section includes an outline of a mathematical formulation of a variety of information systems that refers to several mathematical concepts and their algebraic theory (not explained here, but in referred sources). As such, it can be omitted without any loss of understanding of the general idea presented in the next four paragraphs.

The definition of information used here is very general [Schroeder, 2011c, 2022]. It does not refer to any other definable concepts but exclusively to one categorial opposition of one and many. This generality that someone could object to as being excessive, is intentional. Since the concept of information appears in virtually all possible contexts and has as a particular instance the main tool of any intellectual inquiry in the form of language, any other more specific and less general concept would lose some important applications.

There is a natural question about why not go further and simply consider information a category, an undefinable (primitive) concept characterized by axioms. The reason is that this would obscure some unquestionable features of information present in all its contexts. Also, the fundamental features of the one-many opposition can be derived from the immense body of philosophical reflection on it in the diverse philosophical traditions of several civilizations. These features direct the methods for differentiation of a variety of different types of information. Moreover, the rich philosophical tradition of the one-many opposition creates a valuable intellectual environment for the study of information that cannot be replaced by even a long list of axioms. Finally, the opposition influenced the development of set theory, and mathematical models of information are formulated in the language of set theory.

Thus, information is understood as an identification of a variety understood as a resolution of the categorial opposition of one and many. The study of information is focused on this resolution as a transition from many to one. This transition can be achieved by a selection of one out of many (selective manifestation of information), or by the identification of a structure binding this multitude into one (structural manifestation of information). The two manifestations are always present together.

This definition, or rather any definition of information becomes meaningful only when followed by a theory providing theoretical tools for the study of information. For instance, the statements of the theory should have consequences that can be empirically tested. Moreover, they should be consistent with already accumulated results of inquiries of information in the more specific domains of its applications. On the other hand, the concept of information has so large variety of applications that its theory requires a high level of abstraction, and therefore a mathematical formalism.

## 8.2. *Outline of a Mathematical Formalism of the Theory of Information*

Information has multiple contexts and each of them requires a specification of its manifestation in terms of an information system. An information system specifying the type of information in mathematical terms is in this approach a closure space and all mathematical concepts mentioned here for modeling information are expressed in terms of the theory of such spaces [Birkhoff, 1967]. In this study, there is no need to restrict this concept by additional conditions unless we proceed to its application in one of its specific domains. Thus, information can be of the geometric, topological, logical, or physical type associated with additional defining conditions for an appropriate closure space describing an information system. However, here we want to consider a general formalism. This is the reason for using the concept of a general closure space which generalizes formalisms of all these (geometric, topological, logical, or physical theories) and many more mathematical theories [Birkhoff, 1967].

The logic of such a general information system is a complete lattice of closed subsets of the information system. At this point, it is important to indicate that the term "logic" as used here has a much more general meaning than usual which only in the case of linguistic or probabilistic information systems can be identified with the familiar Boolean lattice defined by the connectives between propositions of some language. The less conventional application of this term consistent with the approach of this paper can be found in quantum logics defined on the closed subspaces of a Hilbert space (alternatively, on the projectors on these subspaces) or in Rota's logics of information identified with lattices of partitions.

The instances of the information within an information system are filters (sometimes called dual ideals) defined on the lattice of closed subsets (the logic of the information system). Filters are collections of closed subsets selected from the logic of the information system, such that with each closed subset all closed subsets including it belong to the filter too (i.e. filters are hereditary), and which are closed with respect to finite intersections. Filter in the logic of information systems is a direct generalization of Mill's connotation in the traditional Boolean logic. Ultrafilters, principal filters, and prime filters characterize special types of information. Since filters representing instances of the information are defined on logics that are not necessarily Boolean lattices, and the theory of filters is typically studied in this particular context (e.g. Stone Theorem) it is important to be aware of the ramifications of the theory of filters when we transcend the Boolean context. For instance, ultrafilters are not necessarily prime filters anymore, which is a fact that frequently confuses physicists in discussing the question of hidden variables.

The formalism based on filters reflects structural characteristics of information and filters represent a state of some universe of inquiry (quite frequently simply called "possible worlds"). Certainly, the universe of inquiry should not be confused with the world as the filters can be defined only after a particular logic of information is chosen, i.e. after the type of information is established.

Information defined or characterized as filters can be identified as analog type as they constitute the connotation of information characterizing the state of the inquired system. However, we have an alternative tool for inquiry of information referring to observed numerical characteristics associated with digital type. Here we have σ-fields of subsets and measures defined on them. The measures can be arbitrary, associated with magnitudes characterizing the objects of inquiry, possibly restricted to probability type, or further restricted to binary logical valuations. In each case, we can construct a corresponding lattice that can be interpreted as a logic of information. Furthermore, we can distinguish filters representing instances of information.

Finally, we can establish the relationships between the analog and digital descriptions of information in different types of information systems classified by their logics, i.e. lattices of closed subsets. These relationships are complex and they heavily depend on the specifics of the systems. In general, the best-known and simplest relationships in the Boolean type of information systems become more complex and ramified when their logics are unconventional, i.e. the lattices of closed subsets differ substantially from Boolean algebras, for instance in quantum logics. Then, the logic of qubits is different from the logic of qudits.

The explanatory power of the formulation of information theory in terms of closure spaces can be appreciated even more when we recall that the famous construction of real numbers in terms of Dedekind cuts is nothing else but the identification of real numbers with the closed subsets of the set Q of rational numbers with respect to the Galois closure (polarity) defined by the order relation of Q. This makes it possible to interpret magnitudes and measures as structure-preserving functions (morphisms) between the lattices of closed subsets (logics) and to get better insight into the role of real numbers in the study of information.

Further details of this theory of information can be found elsewhere in my already published papers [2019, 2022] while additional details and demonstrations or proofs of the mathematical claims made here will appear in a paper currently in preparation.

## 9. Conclusions

The paper demonstrates the urgent need for further intensive studies of information as the only tool for securing human control of rapidly developing information technologies. Legal regulations may restrict or direct human actions, but even if effective in such tasks (which is doubtful) they will not prevent the dangers of unpredictable developments in technology.

The further development of the study of information has to be coordinated with the studies of related important but poorly understood concepts of consciousness, complexity, intelligence, computation, and life (in this article the first three are considered). The wide range of phenomena involving these concepts requires that adequate theories of information have to be at a sufficiently high level of abstraction. On the other hand, they have to be sufficiently specific and precise in their methodologies to have explanatory power for the entire complex of studies not only of information but also of consciousness, complexity, and intelligence.

An outstanding deficiency in the study of information at a high level of generality is the negligence of the semantics of information which if considered at all is typically formulated in restricted contexts, such as the context of life in biosemiotics. All existing semantic theories of information have in their center a triangular relation relativizing the meaning to non-informative elements such as an interpreter, thought, etc. A binary approach excluding the mediation of a third party is proposed.

The most important epistemological obstacles identified in this paper have methodological character. They are related to misunderstandings of the traditional divisions into quantitative vs. qualitative methodologies, the role of mathematics, in particular, the role of number theory, measure theory, and probability theory. Another source of confusion identified in this paper was a more specific misunderstanding of the distinction between analog and digital information. In each case, some proposals were presented for how to eliminate confusion.

The final part of the paper provided a very brief and general exposition of an example of a theory of information in which the epistemological obstacles are avoided.

Further study in this direction may address a question about generative AI. Can we identify essential differences between generative AI systems and human intelligence? My working hypothesis is that the most important difference is in their information logics (understood in the way described in the preceding section). In the former, the logic of information is built based on Large Language Models (LLM) derived from the characteristics of the language. Human intelligence has its logic reflecting not relations within the language but in the model of reality or world that we develop in our living experience. Thus, to acquire human intelligence the AI systems have to be built not based on the patterns of the language, but the patterns of reality which we could call "Large World Models". The formalism of the theory of information presented above is consistent with this idea.

**Acknowledgment** The author would like to express his gratitude to the anonymous reviewer for helpful suggestions regarding the format of the work.